\documentclass[letterpaper, 10 pt, conference]{ieeeconf}  

\IEEEoverridecommandlockouts                              

\usepackage{color}

\usepackage{times}
\usepackage{epsfig}
\usepackage{graphicx}
\usepackage{tabularx}
\usepackage{amsmath}
\usepackage{amssymb}
\usepackage{bbm}
\usepackage[misc]{ifsym}
\usepackage[export]{adjustbox}
\usepackage{subfigure}

\usepackage[ruled,vlined,linesnumbered]{algorithm2e}
\usepackage{graphicx} 

\usepackage{subfigure}
\usepackage{tikz}
\usepackage{verbatim}
\usepackage{booktabs}
\usepackage{multirow}
\usepackage{makecell}
\usepackage{authblk}
\usepackage{hyphenat} 

\usepackage{pgf}
\usepackage{algpseudocode}
\usetikzlibrary{arrows,automata}

\usepackage{graphbox}

\makeatletter
\newcommand{\removelatexerror}{\let\@latex@error\@gobble}
\makeatother

\makeatletter 
  \newcommand\figcaption{\def\@captype{figure}\caption} 
  \newcommand\tabcaption{\def\@captype{table}\caption} 
\makeatother

\makeatletter
\let\NAT@parse\undefined
\makeatother
\usepackage[colorlinks]{hyperref}  

\title{\LARGE \bf
Empirical Analysis of Bi-directional Wi-Fi Network Performance on Mobile Robots in Indoor Environments
}

\author{Pranav Pandey \and Ramviyas Parasuraman \thanks{$^{*}$ The authors are with the Heterogeneous Robotics Research Lab, Department of Computer Science, University of Georgia, Athens, GA 30602, USA. 
Email: {\it \{pranav.pandey,ramviyas\}@uga.edu}. }}

\begin{document}

\newtheorem{definition}{Definition}
\newtheorem{theorem}{Theorem}
\newtheorem{lemma}{Lemma}
\newtheorem{proposition}{Proposition}
\newtheorem{property}{Property}
\newtheorem{observation}{Observation}
\newtheorem{corollary}{Corollary}

\maketitle

\begin{abstract}
This paper proposes a framework to measure the important metrics (throughput, delay, packet retransmits, signal strength, etc.) to determine Wi-Fi network performance of mobile robots supported by the Robot Operating Systems (ROS) middleware. We analyze the bidirectional network performance of mobile robots through an experimental setup in an indoor environment, where a mobile robot is communicating vital sensor data such as video streaming from the camera(s) and LiDAR scan values to a command station while it navigates an indoor environment through teleoperated velocity commands received from the command station. The experiments evaluate the performance under 2.4 GHz and 5 GHz channels with different placement of Access Points (AP) with up to two network devices on each side.  The framework is generalizable to vehicular network evaluation and the discussions and insights from this study apply to the field robotics community, where the wireless network plays a key role in enabling the success of robotic missions in real-world environments. 
\end{abstract}

\section{Introduction}
\label{Sec:introduction}

The potential applications of mobile robots and unmanned ground vehicles (UGV) are increasing every day due to heavy research invested in making them more efficient and successful \cite{murphy2016disaster,yoshida2014}. 
Wireless networking plays a crucial role in many robotic missions such as urban search and rescue (USAR) and survey operations, where mobile robots are deployed in indoor or field environments \cite{muralidharan2017first,queralta2020collaborative}.  Similarly, wireless communication plays a key aspect of connected and autonomous vehicles\footnote{Throughout this paper, we use the term "robot" to represent both mobile robots and autonomous vehicles.}, as well as UAVs.

The key role of a wireless network in robotic networks is to transfer valuable data between the robots and the base station. 
We can classify this data based on the exigency (or priority) \cite{tardioli2019pound}. For example: if robots are being used to carry something in which they must maintain a formation, then their position should be communicated to other robots continuously, so their position data is critical. In another example, if robots are being used to survey an area then the data they are collecting from the survey is the critical data and their position, battery life or any other data is useful but not critical.
Communication loss and sub-optimal performance can be costly and can restrict many robot tasks, where data shared between multiple stations or devices are crucial \cite{amigoni2019online}. 
Similarly, the data transfer needs would differ based on the specific tasks, where a balance of throughput and message latency is important \cite{Parasuraman2013,owen2013haptic}.

\begin{figure}[t]
    \centering
    \includegraphics[width=0.95\columnwidth]{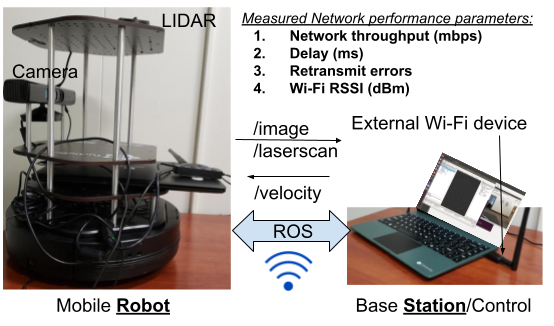}
    \vspace{-4mm}
    \caption{Overview of the network analysis experiment setup.}
    \label{fig:overview}
    \vspace{-6mm}
\end{figure}

Moreover, wireless network performance on mobile robots connected to a base station or to another mobile robot (in a multi-robot system, for example) depends on various metrics such as the radio signal strength variations (hardware level), message retransmission errors, and latency (network level), and the data throughput (application level). 
Therefore, to achieve the best performance, it is necessary to design, plan, and optimize the network parameters beforehand by utilizing radio channel modeling techniques and predicting the radio signal variations across the usage area. 
Analyzing these metrics under real operations will provide useful insights into optimizing the robot networking design and setup such that stable and high-performance connectivity with the robot can be maintained during a mission \cite{caccamo2017rcamp,flushing2014spatial,nguyen2012using}. 

Researchers focused on improving the network performance through hardware innovations such as intelligent antenna \cite{min2018directional,parasuraman2017new}, algorithmic improvements such as optimized routing \cite{kim2011testbed,das2007mobility}, mobility-based methods such as the use of relay robots \cite{dixon2009maintaining}. However, there is currently no work that provides a comprehensive framework for measuring and analyzing network performance on robots, especially coupled with the Robot Operating Systems (ROS) \cite{quigley2009ros}, which is a widely-used software development middleware framework in the robotics research community.
To address these gaps, we make the following \textbf{contributions} in this paper.
\begin{itemize}
    \item We contribute an open-source ROS package\footnote{\url{https://github.com/herolab-uga/ros-network-analysis}} to measure important network parameters such as data throughput, round trip delay (latency), Wi-Fi received signal strength indicator (RSSI), and network errors.
    \item To demonstrate the utility of the proposed network measurement framework, we analyze the bidirectional wireless network performance using the proposed framework by conducting extensive empirical experiments, where a mobile robot follows an exploration trajectory (as shown in Fig.~\ref{fig:layout}) while connected to a base station through different Wi-Fi channel (2.4 GHz or 5.8 GHz) in an indoor environment under different Access Point (AP) placement (on the robot or the station). 
    \item Specifically, we test our hypothesis that having the AP on the mobile robot results in better network performance and stability, compared to having the AP at the base station with an ad-hoc Wi-Fi setup.
\end{itemize}

These experiments help validate the measurements' framework, and the results provide valuable data to the community in choosing an appropriate Wi-Fi network channel and setup for mobile robot missions.

\section{Literature Review}
\label{sec:literaturereview}

A good deal of research work laid foundations to analyze, optimize, and apply wireless networks for mobile robots and mobile users in general. Some proposed algorithms for choosing the optimum channel and location of AP to receive the best connection quality \cite{parasuraman2013spatial}, while others relied on wardriving-like techniques to search for a better wireless access point to connect to. For instance, in \cite{1181869}, the AP position is optimized along with channel assignment. They have first analyzed the physical space with all the barriers, then selected the candidate locations that can be used to place an AP and chose the best candidate based on the RSSI and channel interference. 
In \cite{jian2018toward}, a self-positioning AP system performed better than Wi-Fi with no AP mobility.
Authors in \cite{8903880} proposed a software solution by using the floor plan of the building and found the best AP locations by predicting the RSSI variations that considered signal decay due to walls and obstacles in that environment. 

Researchers also looked into the network issues from the perspective of robot control and mobility. 
In \cite{szanto2013performance}, it is shown that having bilateral teleoperation between server and client robots degraded the performance because other channels occupy the bandwidth and the continuous communication between the master and slave robots results in saturation of the communication channel, which makes the teleoperation unusable. 
In \cite{gass2005measurements}, the authors conducted an experiment on motion-based connectivity changes by driving a car at different speeds and measured the network performance. Their results suggested optimum times when packet loss is minimum and throughput is maximum. 

Furthermore, the use of wireless network parameters has been crucial in robot localization. For instance, the authors in \cite{zhang2011improved} developed an algorithm to minimize the RSSI error for wireless networks, which helps the centroid localization algorithm to estimate the best place for an Access Point. Another study \cite{5672458} proposed an RSSI linearization algorithm to estimate the best location for AP with Cramer-Rao Bound (CRB) \cite{1212671} as the baseline. In \cite{wu2015indoor}, the authors have modified the transmission powers of the APs to increase the accuracy of AP localization.

While there are publicly available datasets containing measurements of RSSI on a mobile robot \cite{parasuraman2016crawdad,through_wall_2014}, there is very minimal work done to create a framework to analyze robot network performance in autonomous vehicles. In \cite{masaba2019ros}, the authors proposed a ROS-based framework with which users can simulate realistic communication channels between robots. In \cite{tardioli2019pound}, the authors developed a priority-based topic serialization protocol to reduce delay in multi-hop networking through the ROS framework. They further provided a way to emulate the robot sensor and image data with different priorities in ROS\footnote{\url{https://github.com/dantard/unizar-profiling-ros-pkg}}. In \cite{chen2019performance}, the author analyzed the network performance and quality of service of the ROS2 framework.

However, as can be seen, a comprehensive network measurement framework and empirical experimental analysis of Wi-Fi performance under robot mobility to gather data that can optimize robot network are missing in the literature. This paper addresses this problem.

\section{Network Measurement Framework}
\label{sec:measurement}

We propose a ROS-based unified framework working on a Linux operating system to measure real-time or instantaneous parameters. The below wireless network performance parameters can be measured on a specific interface (communication device) at both the server (base station side) and the client (robot side), and then analyzed. Each parameter is outputted as a ROS message of custom message type that includes all inner (sublevel) metrics of that parameter.

Although the measuring frequency for each parameter can be adjusted, we use a standard 1 HZ rate for all metrics, except RSSI, which is measured at a 10 Hz rate. Note, we omit the measurement of packet loss, which is not a real-time measurement of channel quality as it requires a reasonable amount of time (a few seconds) and a static environment (so it does not apply to a mobile robot context).

\subsection{Received Signal Power Measurement}
\textbf{Received Signal Strength Indicator (RSSI)} is measured in dBm units through the system level statistics as $/proc/net/wireless$ provided by the device driver for that interface. RSSI values provide an indication of environment complexity and the distance between the base station and the robot. A high RSSI value (closer to $-20dBm$) means a stable and good connection, while a low RSSI value (closer to $-80 dBm$) means a poor connection that could be lost anytime. The RSSI measurements are crucial to several robot-aided applications, including robot and AP localization, and positioning the robot for improving connectivity \cite{parasuraman2013spatial,min2018directional}.

Let node $i$ and $j$ are neighbor nodes and are part of an already established route. 
In two-way ground-reflection model of wave propagation for free-space propagation path loss, the relation between the received signal power $P_r$ at node $j$ from node $i$ and transmitter–receiver separation distance $d_{ij}$ is defined as follows:
\begin{equation} \label{eq:1}
P_r \approx \frac{P_tG_tG_rh_t^2h_r^2}{d_{ij}^{4}} ,
\end{equation}
where $P_t$ is the transmitted power, $G_t$ and $G_r$ are gains of transmitting and receiving omnidirectional antennas, $h_t$ and $h_r$ are transmitter and receiver antenna heights and $d_{ij}$ is the distance between transmitting $ith$ and receiving $jth$ nodes. 

In real-world measurements, the signal strength attenuation depends not only on the distance (path loss), but also on the environmental factors such as the objects in the environment (shadowing) and spatio-temporal dynamics (multipath fading) \cite{Lindhe2007}. Therefore, converting the signal power from Eq.~\eqref{eq:1} into RSSI dBm units, we have
\begin{equation}
\begin{split}
RSSI (dBm) & = \underbrace{RSS_{d_0} - 10\eta\log_{10}(\frac{d}{d_0})}_{path\;loss} \label{eqn:elnsm} \\
& - \underbrace{\Psi_{(d_{ij})}}_{shadowing} - \underbrace{\Omega_{(d,t)}}_{multipath} .
\end{split}
\end{equation}
Here, $RSS_{d_0}$ is the RSS at a reference distance $d_0$ (usually 1 m), which depends on the transmit power, antenna gain, and the radio frequency used. $\eta$ is the path loss exponent, which is a propagation constant of a given environment. $\Psi \sim \mathcal{N}(0,\sigma)$ is a Gaussian random variable typically used to represent shadowing, while $\Omega$ is a Nakagami-distributed variable representing multipath fading. 

\subsection{Throughput Measurement}
\textbf{Throughput} is recorded through the link utilization metrics such as the transmitted and received total packets, TCP-specific segments, and UDP-specific datagrams. This is achieved through capturing the Linux system process-level information from $/proc/net/dev$ for a specific wireless interface. By adding the transmitted and received bytes, we get the total throughput in Mbps. Having a high throughput is desirable as it ensures as many sensors and image data are being transmitted across the network, which could help gain high situation awareness to the human operator commanding the robot from the base station \cite{georg2020longtime}.

According to the Shannon–Hartley capacity theorem, in a wireless channel $C$, the maximum
communication channel capacity $C_c$ and the data throughput $T_c$ is related to the RSSI as follows \cite{parasuraman2013spatial}:
\begin{equation} 
    C_c  = B\log_2 (1+10^{((RSSI_{dBm}-PN_{dBm})/10)}) \label{eq:2}
\end{equation}
\begin{equation} \label{eq:3}
T_c = \alpha\textsubscript{T}C_c    
\end{equation}
Here, B is the frequency bandwidth of the channel in MHz, and $PN$ is the background signal (noise) power in the channel. The data throughput $T_c$ is a factor (by
$\alpha\textsubscript{T}$ ) of the maximum capacity $C_c$. From equation \ref{eq:2} and \ref{eq:3} :
\begin{equation} \label{eq:4}
T_c = \alpha\textsubscript{T}B\log_2 (1+10^{((RSS_{dBm}-PN_{dBm})/10)})    
\end{equation}
In our measurements, we measure the actual throughput $T_c$, which mainly varies based on the distance between the nodes and the errors in the network. 

\subsection{Application-level Network Delay Measurement}
\textbf{Network Delay} The network latency is measured in milliseconds as the round trip time delay at the ROS application level. This is achieved through the use of ROS ActionLib, where an action server runs at the base station and the action client runs on the mobile robot. The role can be reversed. A default value of $-1$ in the measurement indicate a loss in connection or a timeout to reach the server from the client. Any active connection will have a delay between 1ms and 2000ms. Keeping the latency (delay) low helps the robot meet high-rate control requirements \cite{Parasuraman2013}.

Theoretically, we know that there are four different types of delays in a network: Transmission Delay; Propagation Delay; Queuing Delay; and Processing Delay. Except the transmission delay, all the other types are not affected by the mobility of the nodes, so we are focusing on study the transmission delay. Now, we also know that transmission delay is the time required to put an entire packet into the communication media. It can be computed by the following equation. 
\begin{equation} \label{eq:5}
    Delay = L/R
\end{equation}
Here, L is the length of a packet and $R$ is the transmission rate. Considering the transmission rate as the actual throughput of the network,  we can replace $R$ with $T_c$. 
\begin{equation} \label{eq:6}
    Delay = \frac{L}{\alpha\textsubscript{T}B\log_2 (1+10^{((RSS_{dBm}-PN_{dBm})/10)})}
\end{equation}

\subsection{Driver-level Network Errors Measurement}
\textbf{Network Errors} Network errors are metrics such as an error in transmitting messages (retransmitted packets), errors in receiving (dropped) messages, etc. This is measured through the $netstat$ tool, which is useful for checking the network configuration and activity generic to the computer (not to the interface). For wired interfaces, we also get statistics from the $ethtool$, which queries the hardware device driver. Network errors indicate how good is the wireless channel in dealing with interferences and mobility. Having fewer network errors is desirable for healthy robot communication. Currently, there are no specific theoretical relationship available between the signal power and the various network errors, since they are dependent on a lot of factors affecting the network channel and the processing at the device.

\section{Experimental Analysis}
\label{sec:expreiment}

Many factors affect network performance. For instance, multiple devices using the same network could negatively impact the performance, while the positioning of the devices could improve the performance. Some other factors such as the transmission power, interfering devices are not in control of developers. 
Motivated by these, we look at analyzing the performance impacted by the following factors: devices on the robot and base station (internal or external interfaces), AP placement (Router, AP on the robot, AP on the base station), frequency channel used by the network (2.4/5 GHz).

\subsection{Experiment Setup}
\label{sec:expsetup}

\begin{figure}[t]
    \centering
    \includegraphics[width=\columnwidth]{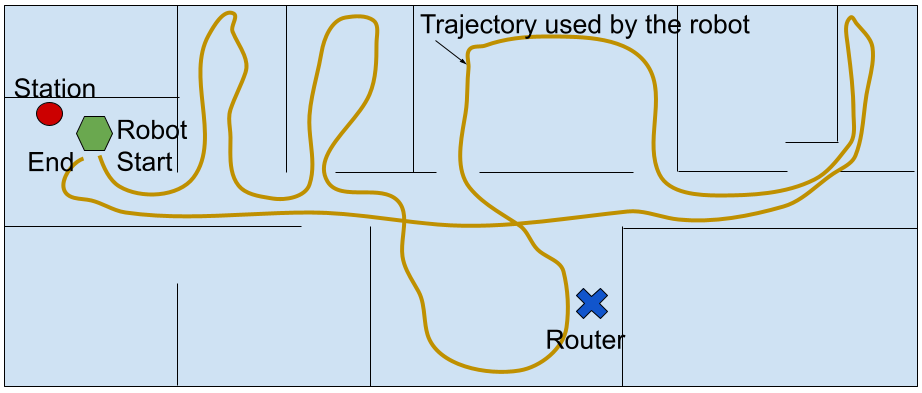}
    \vspace{-6mm}
    \caption{Experiment setup showing the position of the stations and the trajectory of the robot used to simulate a realistic robot exploration task.}
    \label{fig:layout}
    \vspace{-4mm}
\end{figure}
On the robot side, we use a turtlebot mobile robot platform equipped with a laptop, a Kinect RGB-D camera, a 2D laser scanner, and an internal (iface 1) and an external (iface 2) Wi-Fi adapter. On the station side, we use a laptop with an internal (iface 3) and an external (iface 4) Wi-Fi adapter. All Wi-Fi adapters support IEEE 802.11ac protocol with dual-band (2.4 GHz and 5.8 GHz) frequency channels. The external adapters have high-gain antennas compared to the internal ones. In addition, we also use a Wi-Fi router to which the robot and station are connected as the baseline setup.

Both robot and station use ROS as the middleware framework. The mobile robot sends the raw camera image data (RGB) and the LIDAR scan data to the station, where we view these data through the \textit{rviz} and \textit{rqt} tools of the ROS visualization stack. 
The \textit{roscore} (master program node) that establishes connection within the ROS machines is running on the robot through the first interface, i.e. the internal Wi-Fi interface of the laptop. 

Fig.~\ref{fig:layout} shows the indoor environment floor plan and the trajectory made by the mobile robot during the experiments through manual teleoperation. The command station is always stationary. We performed two trials each and report the average. The exact replication of trajectory do not matter to these experiments, but for comparative analysis between several experiment cases a rough following of trajectory has been manually verified through the recorded data.

\subsection{Experiment Cases}
\label{sec:expcases}

For empirical analysis, we experiment with 10 different cases (shown in Table~\ref{tab:cases}) under which we control the mobile robot to trace the same trajectory while measuring the network performance through the proposed framework in Sec.~\ref{sec:measurement}. Case 1 is the baseline for 2GHz connections and case 2 is the baseline for 5GHz connections. Both the baseline cases use a stationary Wi-Fi router as an intermediary for their networking. The cases where we mention "Direct" are the ones in which the robot or base station acts as the AP and the other side connects to it (without a router). That is, one out of the four interfaces act as an AP, while the other interfaces communicate through this AP.

\begin{table}[]
\caption{Experiment cases used to analyze the network performance.}
\label{tab:cases}
\vspace{-2mm}
\centering
\resizebox{\columnwidth}{!}{
\begin{tabular}{ |c|c|c|c| }
\hline
\multicolumn{1}{|c|}{} & \multicolumn{1}{|c|}{} & \multicolumn{2}{|c|}{Interface Used}\\	
\hline
\multicolumn{1}{|c|}{Case No.} & \multicolumn{1}{|c|}{Direct/Router} & \multicolumn{1}{|c|}{Robot} & \multicolumn{1}{|c|}{Station}\\
\hline
\multicolumn{4}{c}{2 GHz Wi-Fi channel experiments}\\
\hline
Case 1 &	Router &	iface 1 &	iface 3\\
Case 3 &	Direct &	iface 1 (AP) &	iface 3\\
Case 5 &	Direct &	iface 1 &	iface 3 (AP)\\
Case 7 &	Direct &	iface 2 (AP) &	iface 4\\
Case 9 &	Direct &	iface 2 &	iface 4 (AP)\\
\hline
\multicolumn{4}{c}{5 GHz Wi-Fi channel experiments}\\
\hline
Case 2 &	Router &	iface 1 &	iface 3\\
Case 4 &	Direct &	iface 1 (AP) &	iface 3\\
Case 6 &	Direct &	iface 1 &	iface 3 (AP)\\
Case 8 &	Direct &	iface 2 (AP) &	iface 4\\
Case 10 &	Direct &	iface 2 &	iface 4 (AP)\\
\hline
\end{tabular}
}
\vspace{-5mm}
\end{table}

\begin{table*}[t]
\caption{Summary of Network Performance Measurements in each case. M: Mean, std: standard deviation.}
\label{tab:measurements}
\vspace{-2mm}
\centering
\resizebox{\linewidth}{!}{%
\begin{tabular}{ |c|c|c|c|c|c|c|c|c|c|c| }
\hline
\multicolumn{1}{|c|}{Cases} & \multicolumn{2}{|c|}{Throughput (mbps)} & \multicolumn{3}{|c|}{Network Delay (ms)} & \multicolumn{4}{|c|}{RSSI (dBm)} & \multicolumn{1}{|c|}{Retransmitts (\#)}\\	
 	\hline
\multicolumn{1}{|c|}{} & \multicolumn{1}{|c|}{M} & \multicolumn{1}{|c|}{std} & \multicolumn{1}{|c|}{M} & \multicolumn{1}{|c|}{std} & \multicolumn{1}{|c|}{\% loss} & \multicolumn{1}{|c|}{ M (at Robot) }  & \multicolumn{1}{|c|}{std } & \multicolumn{1}{|c|}{ M (at Station) } & \multicolumn{1}{|c|}{ std  } & \multicolumn{1}{|c|}{Cumulative}\\	
\hline
\multicolumn{9}{c}{2 GHz Wi-Fi channel experiments} \\
\hline
Case 1 (Router) &	12.8 &	3.6 & {47.8} &	37.7 & 0 & -57.5 & 7.4 & -54.7 & 1.5 & 2376\\
\hline

Case 3 (Robot-AP) &	19.1 &	4.3 &	163.6 &	270.7 & 9.6 & N/A & N/A & -63.7 &	12.7 & 133\\
\hline

Case 5 (Station-AP) &	13.1 &	5.6 &	\textbf{96.6} &	218.2 & 0.1 & -67.4 &	11.9 & N/A & N/A & \textbf{96}\\
\hline
Case 7 (Robot-AP) &	\textbf{50.2} & 20.3 & 124.8 &	111.7 & 0.6 & -57.9 & 16.0 & -56.5 & 16.5 & 324\\
\hline
Case 9 (Station-AP) &	35.9 & 16.9 & 116.4 &	152.6 & 0.3 & \textbf{-51.4} & 16.5 & \textbf{-44.6} & 14.2 & 235\\
\hline
\multicolumn{9}{c}{5 GHz Wi-Fi channel experiments} \\
\hline
Case 2 (Router) &	{73.1} & 13.9 & {48.2} &	50.4 & 0 & -58.8 &	7.0 & -60.6 & 1.4 & 273\\
\hline
Case 4 (Robot-AP) &	21.6 &	8.5 &	184.1 &	264.8 & 11.9 & N/A & N/A & -62.6 &	13.6 & 688\\
\hline

Case 6 (Station-AP) &	16.5 &	6.5 &	\textbf{62.9} &	62.4 & 0 & -62.0 &	13.7 & N/A & N/A & 402\\
\hline

Case 8 (Robot-AP) &	\textbf{68.6} & 23.8 & 72.9 &	132.1 & 0 & \textbf{-57.0} & 17.1 & -55.0 & 16.9 & 2210\\
\hline

Case 10 (Station-AP) &	66.4 & 24.6 & 109.1 &	167.4 & 0 & -69.9 & 25.1 & \textbf{-46.8} & 11.9 & \textbf{126}\\
\hline
\end{tabular}
}
\vspace{-4mm}
\end{table*}

\subsection{Experiment Hypothesis}

From the equations Eq.~\eqref{eqn:elnsm}, \eqref{eq:4}, and \eqref{eq:6}, we have the relationships between throughput, delay, signal power, and the mobility of the nodes (changes in the distance). Based on these relationships, we can say that if during the mobility, the distance between the nodes $d$ increases then the RSSI will decrease and the delay will increase, while the throughput will decrease. Theoretically, these relationships should be maintained for both 2GHz and 5GHz channels. However, it is not easy to obtain theoretically how the mobility impacts these parameters when the Access Point is on one of the nodes (fixed station or on the mobile robot). Through these experiments, we aim to test what is the best way to increase the throughput while keeping the delay as low as possible. 

Our hypothesis is that having an Access Point on the vehicle (robot) is better than connecting the robot to an AP at a fixed station mainly due to the following reasons: 1) AP on the robot enables network interface management easier and creates a potential to apply machine learning algorithms on the robot eventually optimizing the wireless network; 2) mobility of the robot can limit the impact of multipath fading thereby obtaining higher RSSI during the movement \cite{parasuraman2013spatial}.
 
\section{Results and Discussion}
\label{sec:results}

\begin{figure}[t]
    \centering
    \includegraphics[width=0.9\columnwidth]{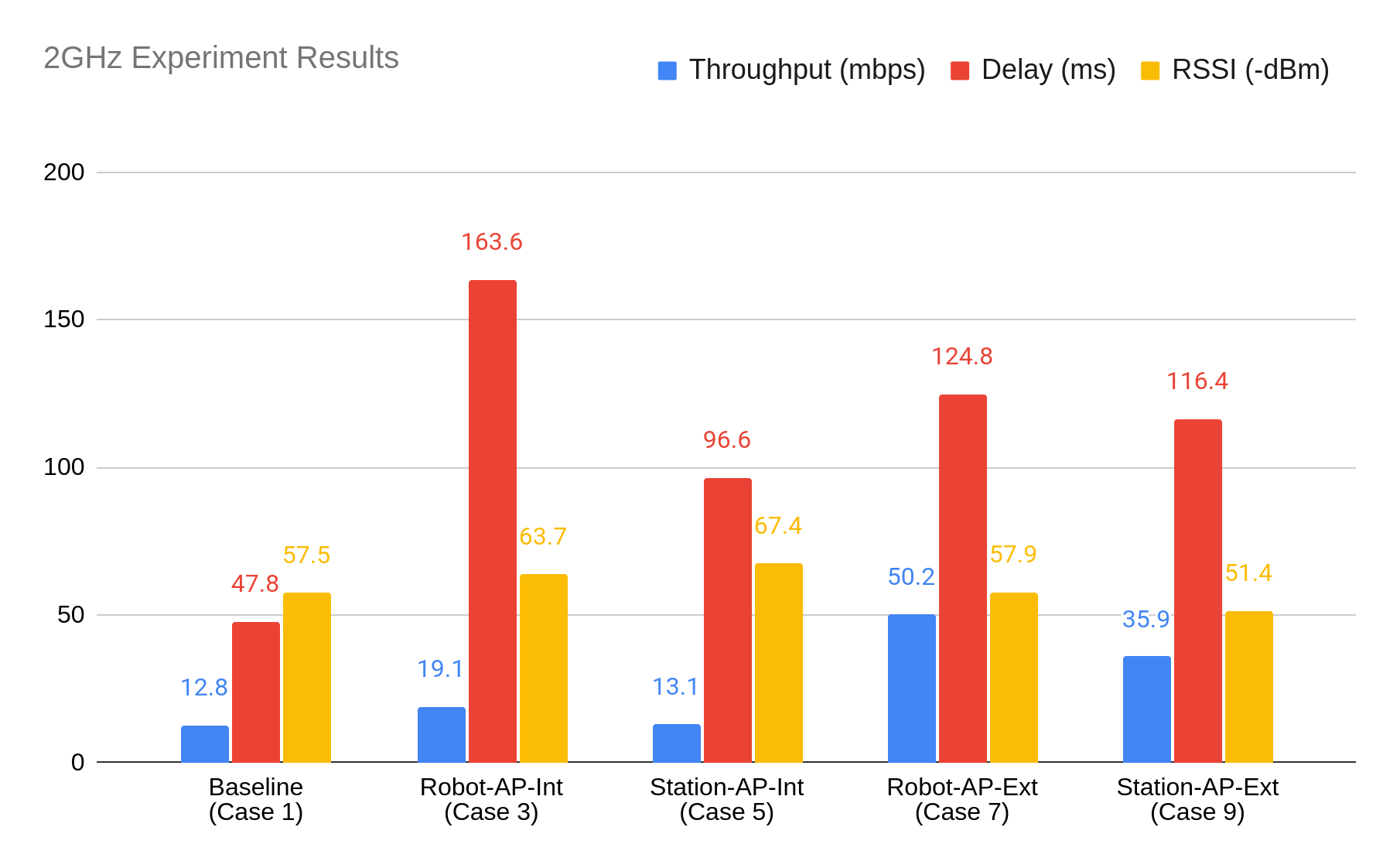}
    \vspace{-4mm}
    \caption{Measurement for cases with 2 GHz channel.}
    \label{fig:mesurement-2ghz-oneiface}
    \vspace{-4mm}
\end{figure}

\begin{figure}[t]
    \centering
    \includegraphics[width=0.9\columnwidth]{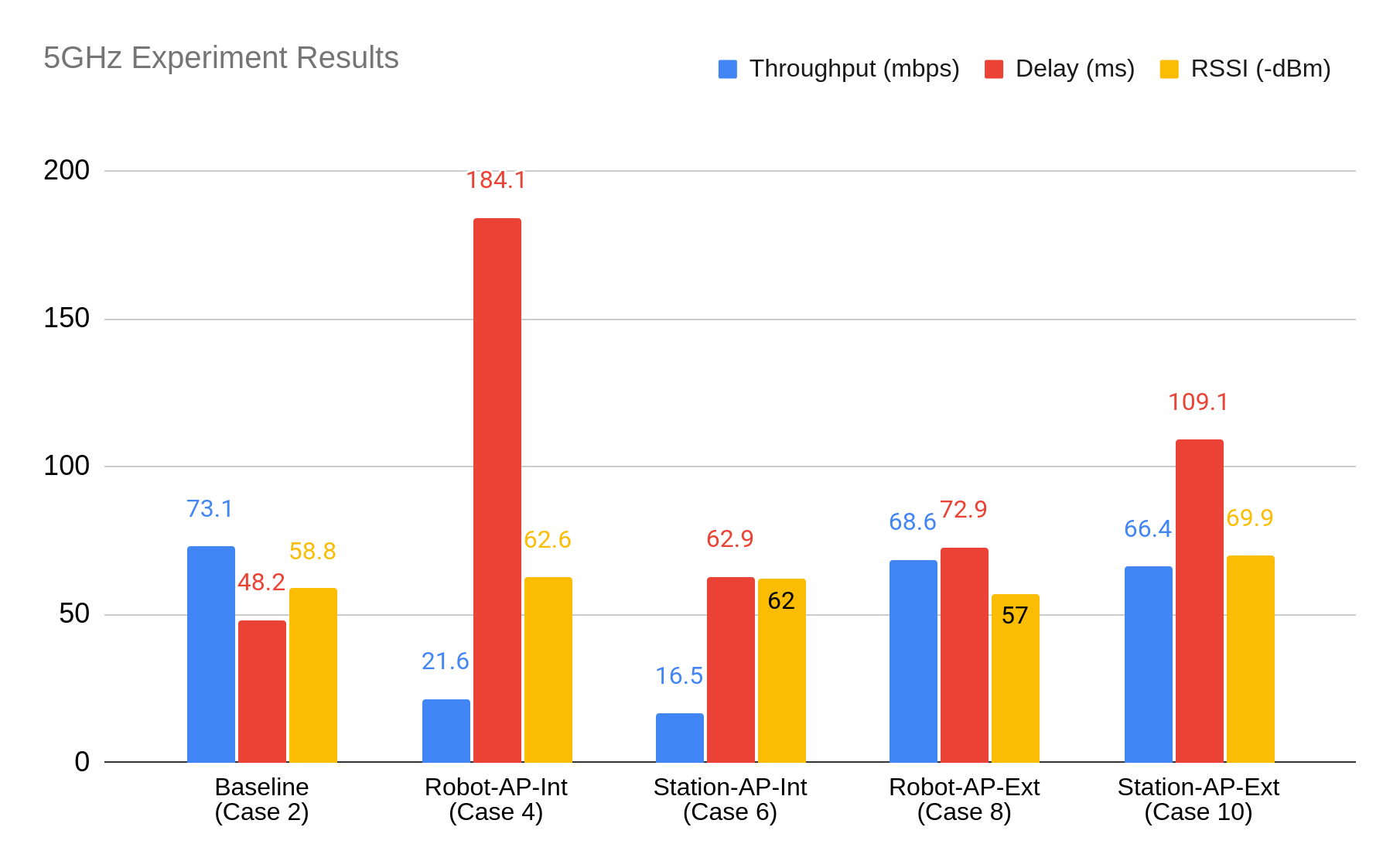}
    \vspace{-4mm}
    \caption{Measurement for cases with 5 GHz channel.}
    \label{fig:mesurement-5ghz-oneiface}
    \vspace{-4mm}
\end{figure}

Table~\ref{tab:measurements} provides the measurements with statistics for the throughput, delay, RSSI, and error metrics, respectively. In the table, we highlight the best measures (except baseline) for each metric in bold. 
Fig. ~\ref{fig:mesurement-2ghz-oneiface} and \ref{fig:mesurement-5ghz-oneiface} present a comparative summary of the observed performance metrics in 2.4 GHz and 5.8 GHz channel, respectively.
Further, we recorded a $rosbag$ data of the network performance parameters measured on both sides of the network for each experiment case and trial. We also released these datasets in the same GitHub repo that is hosting our measurement framework.
Below, we analyze the network performance from different channel perspectives. 

In the Table~\ref{tab:measurements} we provided the data for throughput, delay and error metrics measured at the robot side only, but for RSSI value we have provided the data of both sides, when available (e.g., provided by the device driver). 
For instance, the external adapters with device driver rtl88x2bu is giving us advantage of recording the RSSI value of an interface while it act as an AP as well (cases 7 to 10). 
We use robot side metrics in general for our analysis. We always run the $roscore$ on robot which is transmitting the data. While recording these data we observed one very interesting verity that receiving data is more than the transmitting data. After looking for the reason why received data is more than transmitted data we found that if $roscore$ is running on a system which is transmitting the data then the receiving data will always be more than transmitted data, this may be because $roscore$ is publishing some topics like $/rosout$ and $/rosout\_agg$ which is transmitting some data to the other nodes which are not acting as master node.


\subsection{2GHz Channel used}
\label{sec:2one}

\begin{figure*}[htp]
\centering
 \subfigure[{2GHz - Throughput}]{
    \begin{minipage}[t]{0.32\textwidth}
        \includegraphics[width=1.1\linewidth]{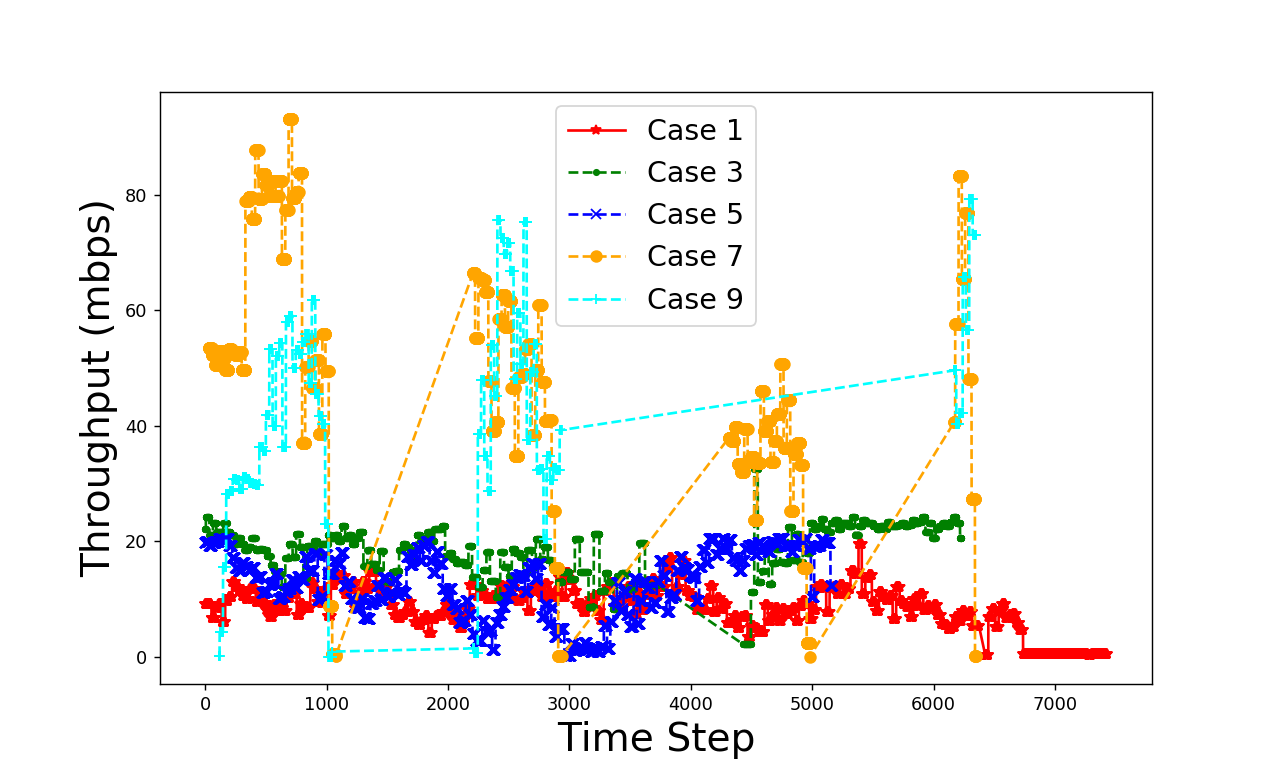}
        \label{fig:throughput-2ghz-oneiface}
        \vspace{-2mm}
    \end{minipage}}
\subfigure[{2GHz - Delay}]{
    \begin{minipage}[t]{0.32\textwidth}
        \includegraphics[width=1.1\linewidth]{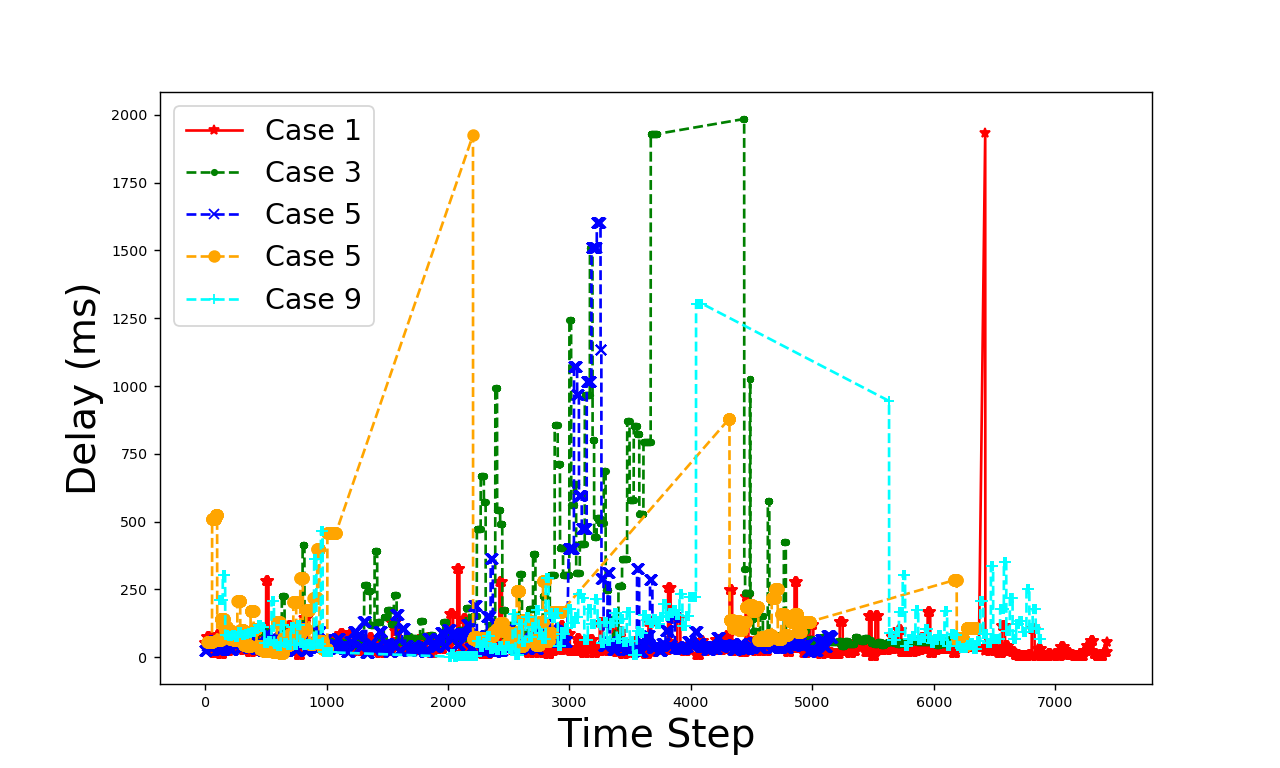}
        \label{fig:delay-2ghz-oneiface}
        \vspace{-2mm}
    \end{minipage}}
    \subfigure[{2GHz - RSSI}]{
    \begin{minipage}[t]{0.32\textwidth}
        \includegraphics[width=1.1\linewidth]{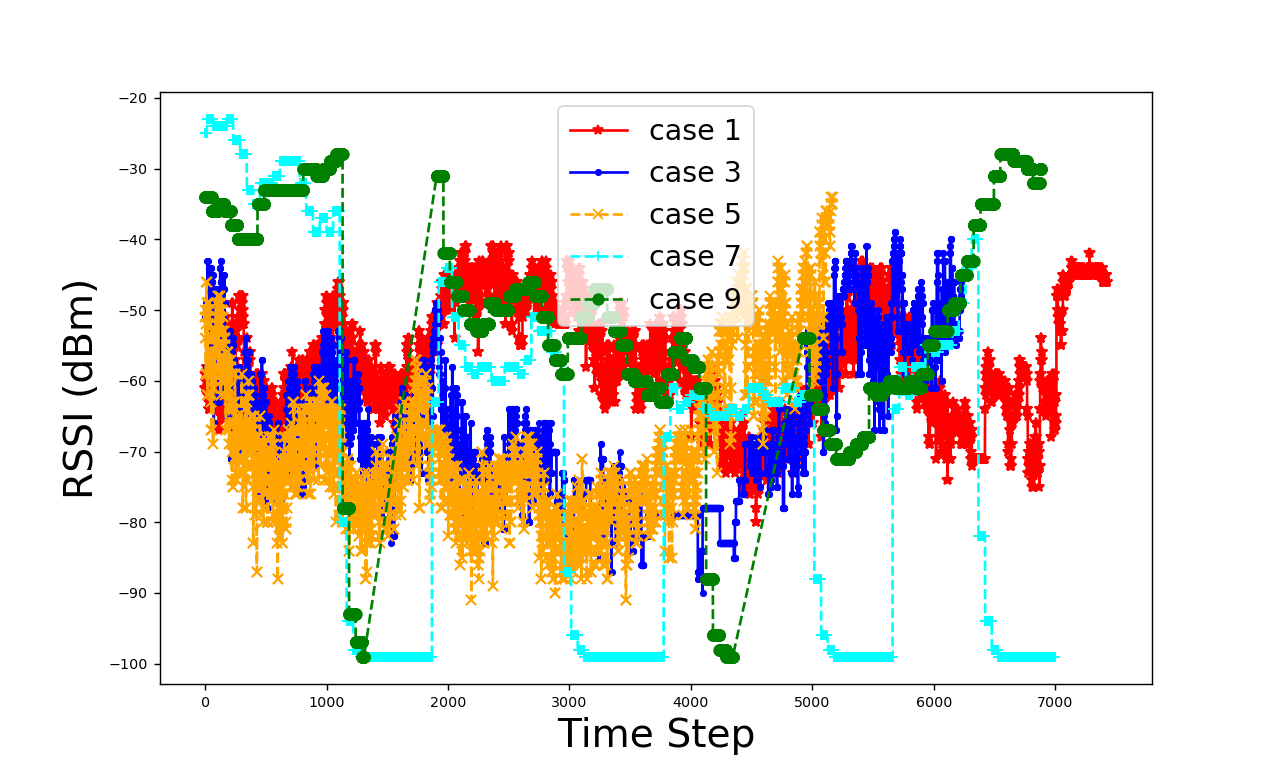}
        \label{fig:rssi-2ghz-oneiface}
        \vspace{-2mm}
    \end{minipage}}
    \vspace{-2mm}
\caption{Instantaneous Throughput and Delay metrics for 2GHz channel experiments.}
\label{fig:throughputanddelayrssi2G}
\vspace{-6mm}
\end{figure*}    
\begin{figure*}[htp]
\centering
 \subfigure[{5GHz - Throughput}]{
    \begin{minipage}[t]{0.32\textwidth}
        \includegraphics[width=1.1\linewidth]{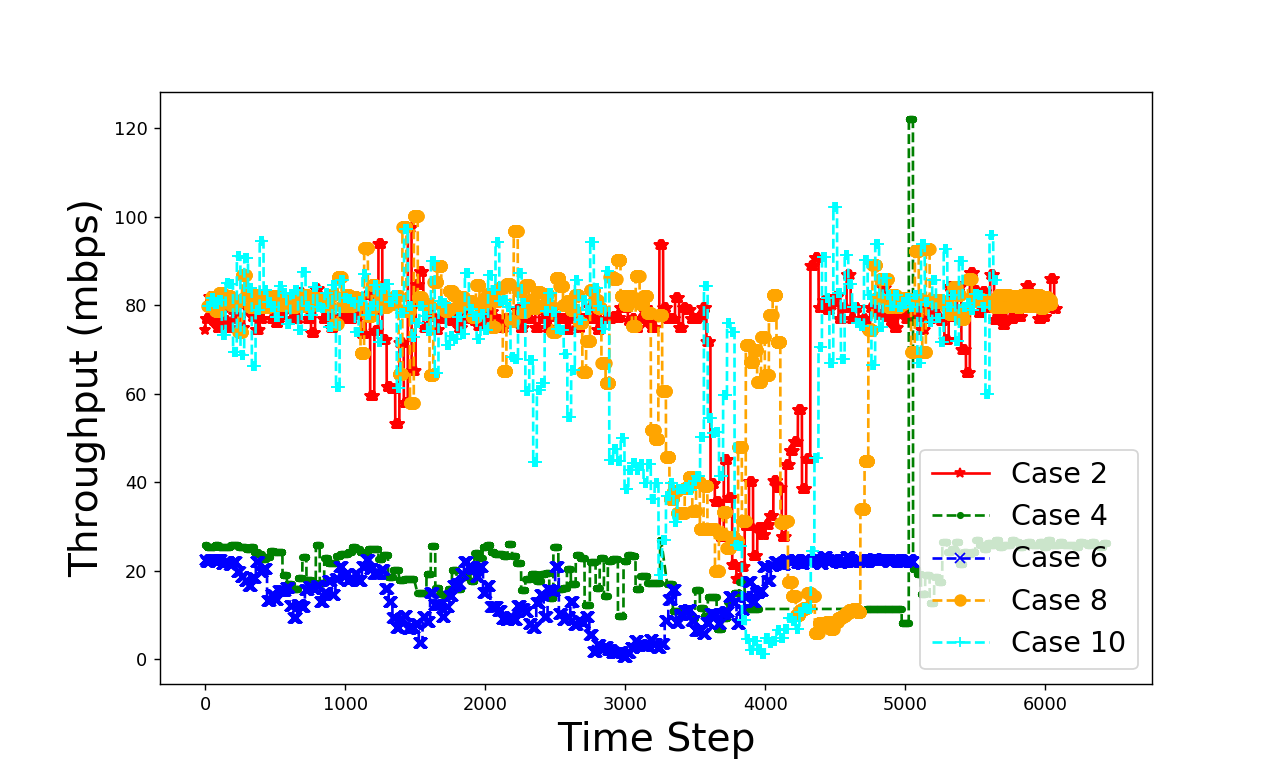}
        \label{fig:throughput-5ghz-oneiface}
        \vspace{-2mm}
    \end{minipage}}
\subfigure[{5GHz - Delay}]{
    \begin{minipage}[t]{0.32\textwidth}
        \includegraphics[width=1.1\linewidth]{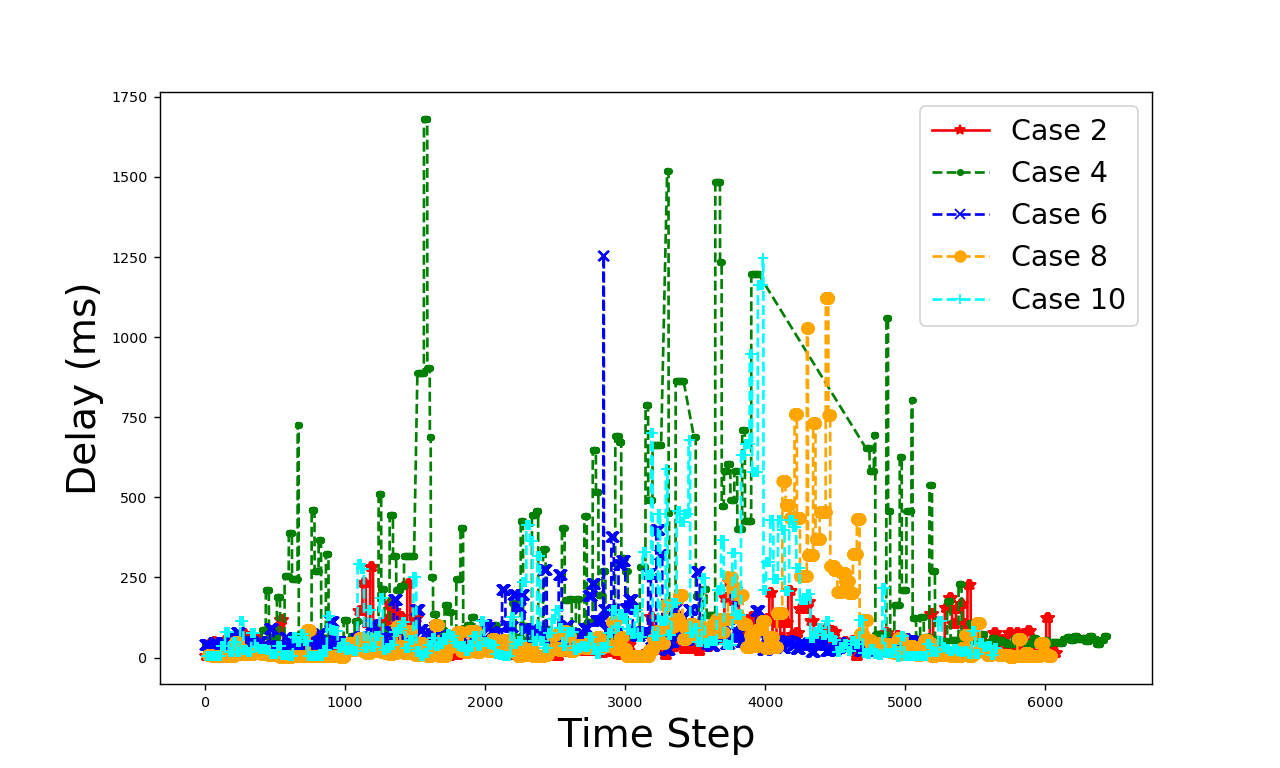}
        \label{fig:delay-5ghz-oneiface}
        \vspace{-2mm}
    \end{minipage}}
    \subfigure[{5GHz - RSSI}]{
    \begin{minipage}[t]{0.32\textwidth}
        \includegraphics[width=1.1\linewidth]{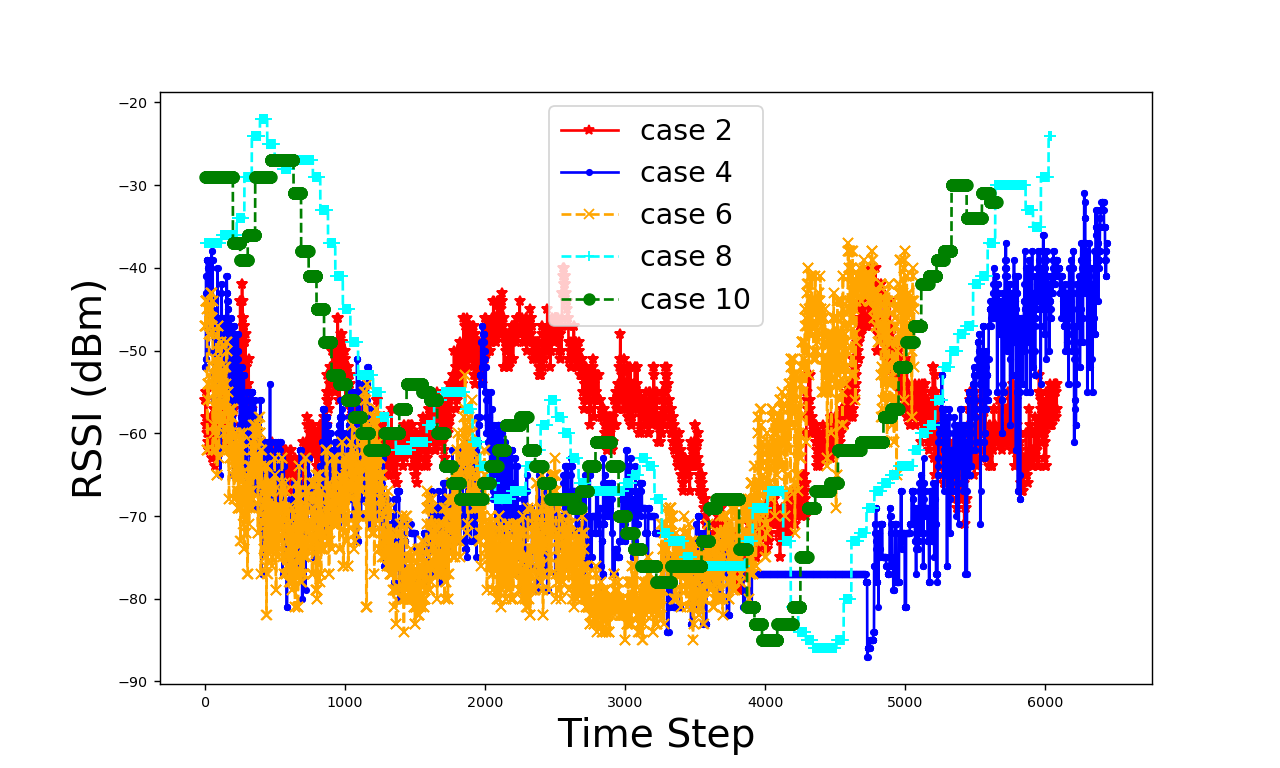}
        \label{fig:rssi-5ghz-oneiface}
        \vspace{-2mm}
    \end{minipage}}
\vspace{-2mm}
\caption{Instantaneous Throughput and Delay metrics for 5GHz channel experiments.}
\label{fig:throughputanddelayrssi5G}
\vspace{-4mm}
\end{figure*} 

Here, we compare and analyze the performance results of the two interfaces used under the 2GHz channel - Case 1 (baseline), Case 3, Case 5, case 7, and Case 9. 
For Case 1, both station and robot are connected to the router, case 3 and case 7 is when the AP is on the robot and station is connected to the AP whereas, in case 5 and case 9 the AP is on station and robot is connected to the AP. Case 3 and case 5 uses internal Wi-Fi adapter, whereas case 7 and case 9 uses external Wi-Fi adapter. 

In terms of throughput, From Fig.~\ref{fig:throughput-2ghz-oneiface} we can see that having an AP on the robot is better than having it on the station as the throughput is higher in case 7 than any other cases. In case 1 (baseline), the low throughput could be because multiple devices are connected to the router during these experiments (we used a lab router to which multiple laptops connect) and therefore, the bandwidth is divided. 

On the other hand, the latency (delay) performance of case 7 is worse than case 1, 3, 5, or 9 (Fig.~\ref{fig:delay-2ghz-oneiface}). This increase in delay in case 7 could be attributed to the mobility of the robot which carries the AP, compared to mild delay surges in Case 5 and 9 when the AP is static at the station. The delay also is highest when the robot is far away from the station (around the center of the time steps).

The number of retransmits is lowest for case 5 and highest for case 1, while balanced for other cases. We calculate the connectivity loss as the percentage of time the delay metric reported a $-1$ value (when the ROS action server cannot be reached or when there is a long delay timeout of more than 2s).
For cases 1, no temporary disconnection is found between AP and client, but for case 3 it was 9.6\% due to the high delay around the farthest portion of the trajectory. For case 5, 7 and 9 there was only 0.1\%, 0.6\% and 0.3\% respectively, which is negligible as our robot is moving. 

In terms of RSSI readings (Fig.~\ref{fig:rssi-2ghz-oneiface}), case 9 had better values compared to other cases. Case 9's values were the highest but close to case 1's values because of a powerful router with higher transmission power.
To summarize, the throughput for case 7 (AP on the robot) is desirable, but it comes at the cost of increase in network delay.

\subsection{5GHz channel used}
\label{sec:5one}

Here, we compare the two interfaces cases under the 5.8 GHz channel: Case 2 (baseline), Case 4, Case 6, Case 8 and Case 10. Case 2 is when both the station and the robot are connected to the router, case 4 and case 8 is when the AP is on the robot and the station is connected to the AP, and case 6 and case 10 is when the AP is on the station and the robot is connected to the AP. Case 4 and case 6 uses internal Wi-Fi adapter, whereas case 8 and case 10 uses external adapter.

The throughput (Fig.~\ref{fig:throughput-5ghz-oneiface}) of case 2 is the highest in this part because of the high transmission power of the router which is  using the 5.8 GHz signals that can penetrate the walls well and achieve higher throughput and RSSI followed by case 8 , case 10, case 4 and case 6 respectively. Among cases, 4 and 6, the throughput of case 4 is better. Also, in case 6 we had a shorter trajectory for the robot to move because it got disconnected after the right-most room in Fig.~\ref{fig:layout}. Moreover, among case 8 and case 10, throughput for case 8 is better than case 10.

In terms of delay (Fig.~\ref{fig:delay-5ghz-oneiface}), the performance is very similar to the 2.4 GHz channel, where case 4 has high delays compared to case 6. But case 10 has higher delay than case 8. For cases 2 , 6, 8 and 10, no temporary disconnection is found between AP and client, but for case 4 it was 11.9\% which is highest among all 10 cases due to the high delay around the farthest portion of the trajectory.

Retransmit errors shows the case 8 has the highest errors compared to other cases, yet the network is stable as there was no temporary disconnection. However, other cases have significantly less retransmits than case 8. 

The RSSI values (Fig.~\ref{fig:rssi-5ghz-oneiface}) for case 8 is better than the other cases and is closer to the case 2 whereas, case 10 (robot side) had the worst RSSI value among these cases. The RSSI value did not show much difference between cases 4 and 6. 
In general, case 4 and case 8 (robot on AP) seems to be a better alternative to case 6 and case 10 in terms of throughput, but, again, at a cost of higher network delay.

\subsection{Impact of mobility}
\label{sec:mobility}
Mobility not only influences the energy consumed by a robot \cite{parasuraman2012energy}, but also affects the wireless network performance. In this section, we compare the performance measurement for all cases when the robot is static versus when the robot is mobile (in Table~\ref{tab:measurements}). We study the data when the applied linear and angular velocity of the robot is zero. Table ~\ref{tab:static_measurements} shows that throughput has increased for each case, whereas latency decreases when the robot is not moving as expected because the robot mobility adds uncertainty to the RSSI, negatively affecting delay. But this effect differed significantly between 2.4GHz and 5GHz channels. In the 5GHz channel, throughput for case 8 and 10 is almost similar, but the delay is much less for case 8 (AP on robot). However, in 2GHz channel, throughput for case 7 (AP on robot) is the highest, but the delay is the lowest for case 1.

\begin{table}[t]
\caption{Network performance when the mobile robot is not moving.}
\label{tab:static_measurements}
\vspace{-2mm}
\centering
\begin{tabular}{ |c|c|c|c|c| }
\hline
\multicolumn{1}{|c|}{Cases} & \multicolumn{2}{|c|}{Throughput (mbps)} & \multicolumn{2}{|c|}{Network Delay (ms)} \\	
 	\hline
\multicolumn{1}{|c|}{} & \multicolumn{1}{|c|}{M} & \multicolumn{1}{|c|}{std} & \multicolumn{1}{|c|}{M} & \multicolumn{1}{|c|}{std} \\	
\hline
\multicolumn{5}{c}{2 GHz Wi-Fi channel experiments} \\
\hline
Case 1 (Router) &	12.2 &	3.7 & \textbf{46.4} &	34.9 \\
\hline

Case 3 (Robot-AP) &	22.6 &	1.2 &	58.7 &	17.9 \\
\hline

Case 5 (Station-AP) &	18.1 &	3.2 &	46.9 &	33.2 \\
\hline
Case 7 (Robot-AP) &	\textbf{60.6} & 18.3 & 106.3 &	105.1 \\
\hline
Case 9 (Station-AP) &	39.4 & 13.8 & 107.6 &	82.8 \\
\hline
\multicolumn{5}{c}{5 GHz Wi-Fi channel experiments} \\
\hline
Case 2 (Router) &	78.5 & 5.0 & 46.5 &	43.0 \\
\hline
Case 4 (Robot-AP) &	24.7 &	2.9 &	73.9 &	65.9  \\
\hline

Case 6 (Station-AP) &	22.0 &	0.9 &	38.4 &	13.8 \\
\hline

Case 8 (Robot-AP) &	80.1 & 5.8 & \textbf{13.1} &	17.2 \\
\hline

Case 10 (Station-AP) &	\textbf{80.7} & 7.7 & 29.0 &	25.2 \\
\hline
\end{tabular}
\vspace{-6mm}
\end{table}

Overall, we can deduce that having AP on robot is better than AP on station from throughput (data transfer) perspective, but, for time-critical data with low bandwidth, we could prefer AP at the station since that results in the lowest network latency. Ideally, having both AP and station interfaces actively available on the robot and the station could create an optimal channel usable by the application based on the application requirements. For example, if the data time-critical such as sending velocity commands then the Station-AP interface can be used, while for data-critical packets such as image transmission, the Robot-AP interface can be used. We will explore this possibility in our future work.

\section{Conclusion}

This paper investigated the Wi-Fi network performance on a mobile robot in indoor environments for communicating vital sensor data such as camera image streaming and LiDAR data to a command station in terms of network throughput, latency, signal strength, network error, and percentage of timeouts. 
We presented a comprehensive framework for measuring network performance parameters under the Robot Operating Systems (ROS) framework. We further conducted extensive experiments to verify the framework by analyzing bidirectional network performance between a robot and a station. 
Results show that the cases with high throughput also came with a cost of high latency.

The experiment results provided valuable insights and supported our hypothesis that having the AP on the robot results in better network throughput compared to when the robot is connected to a fixed AP at the base (command) station, if a high-power router-based (infrastructure) setup is not feasible per application. However, for better network latency performance, it is preferable to keep the AP at the fixed station. The results are more significant for a 2.4GHz channel compared to a 5GHz channel.

\bibliographystyle{IEEEtran}
\bibliography{references,newreferences}

\end{document}